# The implementation of a Deep Recurrent Neural Network Language Model on a Xilinx FPGA

Yufeng Hao
Dept. of Electronic, Electrical and Systems Engineering
University of Birmingham, Edgbaston, Birmingham, B152TE, UK
E-mail: yxh663@student.bham.ac.uk

Steven Quigley
Dept. of Electronic, Electrical and Systems Engineering
University of Birmingham, Edgbaston, Birmingham, B152TE, UK
E-mail:S.F.QUIGLEY@bham.ac.uk

*Abstract*—Recently, FPGA has been increasingly applied to problems such as speech recognition, machine learning, and cloud computation such as the Bing search engine used by Microsoft. This is due to FPGAs' great parallel computation capacity as well as low power consumption compared to general purpose processors. However, these applications mainly focus on large scale FPGA clusters which have an extreme processing power for executing massive matrix or convolution operations but are unsuitable for portable or mobile applications. This paper describes research on single-FPGA platform to explore the applications of FPGAs in these fields. In this project, we design a Deep Recurrent Neural Network (DRNN) Language Model (LM) and implement a hardware accelerator with AXI Stream interface on a PYNQ board which is equipped with a XILINX ZYNQ SOC XC7Z020-1CLG400C. The PYNQ has not only abundant programmable logic resources but also a flexible embedded operation system, which makes it suitable to be applied in the natural language processing field. We design the DRNN language model with Python and Theano, train the model on a CPU platform, and deploy the model on a PYNQ board to validate the model with Jupyter notebook. Meanwhile, we design the hardware accelerator with Overlay, which is a kind of hardware library on PYNQ, and verify the acceleration effect on the PYNQ board. Finally, we have found that the DRNN language model can be deployed on the embedded system smoothly and the Overlay accelerator with AXI Stream interface performs at 20 GOPS processing throughput, which constitutes a 70.5X and 2.75X speed up compared to the work in Ref.[30] and Ref.[31] respectively.

*Index Terms*—Deep Recurrent Neural Network (DRNN), language model, overlay, PYNQ.

## I. INTRODUCTION

In contemporary society, researchers have made a vast number of achievements in the Artificial Intelligence (AI) and Machine Learning (ML) fields. In particular, the success of alpha Go has greatly raised confidence in machine for processing the human-machine interaction field. However, in the Natural Language Processing field, it is still difficult to have machines understand language. This problem seems to be the significant obstacle in progress into the AI age. It has also become one of the most challenging topics in the Natural Language Processing (NLP) field. As the core part of NLP, the Language Model (LM) play an important role in this processing. A successful LM not only improves recognition accuracy, but also speeds up the decoding process. Furthermore, with telecommunication and computer technology development, our smart mobile devices have become increasingly powerful in computing capacity which has led to the mobile internet revolution. This paper will seek to find a way to deploy a LM into a mobile device and verify the results on a prototype machine.

### A. Background

Nowadays, Natural Language Processing (NLP) has become a large interdisciplinary research topic which includes linguistics, probability, statistics, Analogue and Digital Signal Processing, Neural Networks, Machine Learning, Parallel Computing, Embedded Systems, Software and Hardware System Design, and Programming Languages. There are many scenarios such as Machine Translation, Intelligent Navigation, Speech Recognition, and Voice Search, using the NLP techniques. Especially, with the development of smart mobile devices, the requirement for intelligent voice interaction in smart devices becomes increasingly stronger.

Language Model (LM) as the core part of the NLP system, maps the words or characters into sentences in the decoding stage, and eventually obtains a maximum probability sentence which generally is closer to human being's speaking habits. In the past several decades, researchers have used different kinds of theories and strategies to structure the Language Model and generate lots of useful and important technologies such as Gaussian Mixture Models (GMM), Hidden Markov Models (HMM), and N-gram LM. Most of speech recognition system used GMMs as the probability model of characters or words. This model is easy to estimate and train while it recognizes different states very well. But the GMM actually is a shallow layer network and it is inefficient to build a model for nonlinear data set. In 2006, an effective method of training the Deep Belief Network (DBN) [1] was presented by Geoffrey Hinton's team, which solved the gradient vanishing problem in deep neural networks and paved the way to the application of Deep Neural Networks, which usually contain more than two hidden layers. From then on, researchers started to introduce the Deep Neural Network into Language Model and Acoustic Model research. Dr. Li Deng's and Dong Yu's team from Microsoft were the first researchers to develop a Deep Neural Network speech processing system successfully [2]. The speech recognition system based on DNN has a different framework compared to previous speech recognition technologies because DNN can generate high dimensional features through multi-layers features extraction. The DNN model system reduces the error rate by 30% compared to conventional models in continuous speech recognition condition [2].

However, these DNN language or acoustic models have been developed and applied on large machines with powerful computation capacity, and usually require a Graphics Processing Unit (GPU) to accelerate computation. However, embedded systems usually have limited computing units and programming capacity. It is challenging but interesting to implement and deploy a DNN model in an embedded system; it is also meaningful because there are a vast number of mobile devices with embedded systems. If an effective and efficient DNN model can be developed in a small embedded system, this will be a promising application area for embedded systems and DNNs. Some researchers have launched DNN on Embedded System Projects. A TensorFlow-on-Raspberry-pi Project was issued by Sam Abrahams in GitHub.com [3]. In this project, he focused on how to install TensorFlow, which is a very popular Neural Network framework programming tool, in an embedded system such as a Raspberry Pi. The installation is appropriate for other embedded system based on the ARM architecture. Also, Matthew Rubashkin, an engineer from Silicon Valley Data Science, has implemented a TensorFlow Image Recognition on a Raspberry Pi [4], which runs well and is able to recognize different pictures with a high accuracy. Moreover, on GitHub a Xilinx research group published a Binary Neural Network (BNN) project on an FPGA [5], which converts the floating point weights and activations in conventional neural network into binary values. The Neural Network model is purely implemented on an FPGA with High Level System tools. In this way, the size of the model and the computation volume can be reduced dramatically, reaching a level that can be run on embedded system. Also further research work done by Eriko Nurvitadhi, et al. from Accelerator Architecture Lab, Intel Corporation, realizes almost 50x acceleration performance improvement over the baseline CPU [6]. These works have greatly promoted the application of DNN in Embedded Systems.

*B. Objective and Methodology*

Inspired by the progression of NLP and the above Neural Network on Embedded System projects, we try to implement a Deep Recurrent Neural Network (DRNN) Language Model on a PYNQ board, which is equipped with a ZYNQ-7020-1CLG400C and supports Python and Jupyter notebook programming. Moreover, considering that the Field Programmable Gate Array (FPGA) SOC platform has plenty of parallel computation resources in its Programmable Logic (PL) side and the ability to execute flexible programs in its Processing System (PS) side with a 32-bit ARM core, we can utilize the Overlay which is a kind of FPGA hardware library used in PYNQ board to create a hardware accelerator for the DRNN.

We use Python to build the program for data processing and DRNN training and verification. Python is very popular in scientific computation and data processing. It is supported well by most of deep learning frameworks such as TensorFlow, Theano, Caffe, and so on, and there are plenty of software libraries in Python such as Numpy, Scipy, Scikit-learn, and so on, which greatly facilitate data sampling, analysis, and processing.

For terms of module training and verification, we chose the Theano Deep Learning Framework. We compared several Neural Network framework programming tools such as TensorFlow, Theano, Torch7, Caffe. We found that Theano supports Python and runs well under a 32 bits Linux Operation System (OS), whereas TensorFlow is powerful but it is more suitable to run under a 64 bit OS because its SparseTensor has to be defined in a 64 bit data type. Torch7 and Caffe are both powerful and popular DNN programming frameworks, but Torch7 does not support Python and Caffe is also a little large for embedded system applications.

## II. PYNQ Architecture and functions

PYNQ is the abbreviation of Python Productivity for ZYNQ [21]. From the hardware architecture perspective, the core chip of PYNQ is a Xilinx ZYNQ Chip, which is a FPGA SOC platform combining Programmable Logic (PL) with Programmable System (PS) to perform signal sampling, processing, and control. Form the software perspective, integrated with the Python language and other programming libraries, PYNQ makes it convenient to develop embedded systems based on FPGA. The PYNQ board is shown in Fig.1.

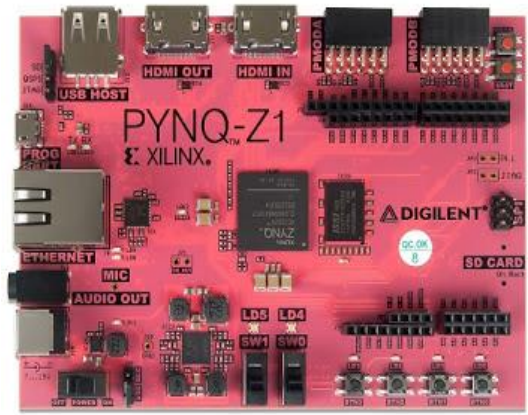

Fig.1. PYNQ board

The ZYNQ chip's internal architecture is shown in Fig. 2. The SOC chip consists of a Dual-Core ARM cortex-A9, a programmable logic part, and other peripheral interfaces such as USB, HDMI, Ethernet, GPIO, and Mic in. The Vivado Design Suite tool can be used to design and configure the PS, PL, and Overlay. An Overlay is a kind of hardware library initiated by Xilinx, which establishes a connection between hardware and software through a Python API. With this connection, it is convenient to exchange data between the software program and logic hardware, which will be helpful to build an accelerator for the algorithm operation. More specifically, because Python is a popular language in natural language processing, machine learning, and artificial intelligence, the PYNQ board is suitable for performing similar applications in these fields.

The implementation of a Deep Recurrent Neural Network Language Model on a Xilinx FPGA 3

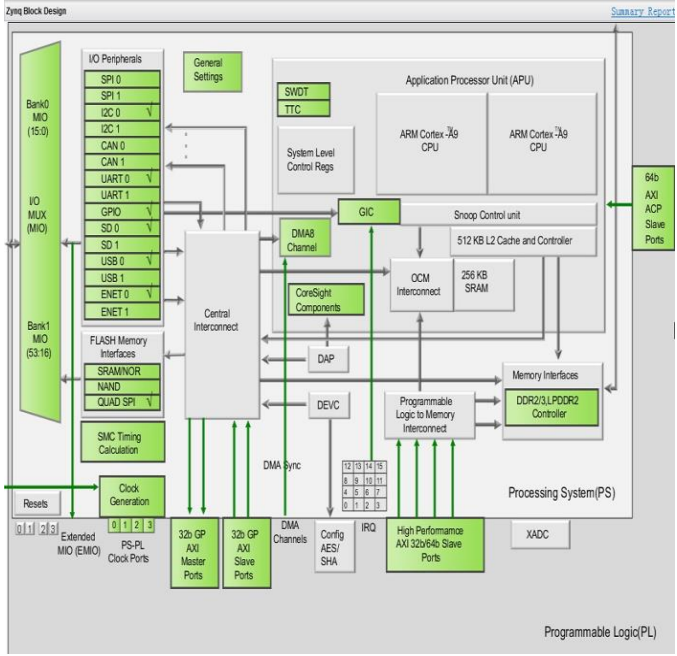

Fig.2. ZYNQ chip internal architecture [7]

## III. Recurrent Neural Network and Language Model

### A. Recurrent Neural Network

An RNN is a kind of neural network that has a memory feature. For general neural networks, from input layer to hidden layer and output layer, the neuron cells in the adjacent layers are fully connected and those in the same layer have no connection [8]. But things are different for the RNN. The hidden layer cells of an RNN look like a combination of general neural network cells in a sequence loop, where one output of one cell will be the input of the next one. That means there exist connections between the neuron cells in the hidden layer of the RNN. These connections give the RNN the capacity to remember previous information. The unit diagram of RNN is shown below.

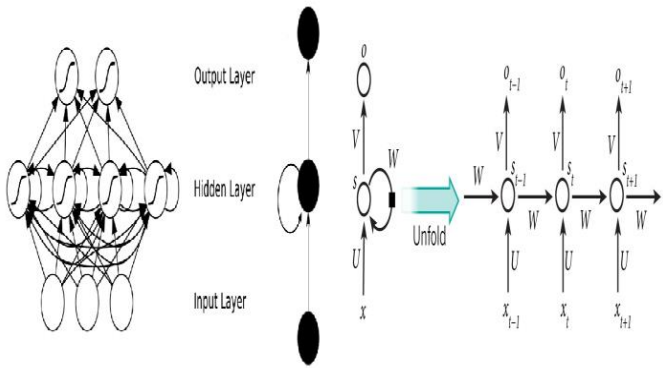

Fig.3. RNN unit perspective modified from [9]

The relationship between the symbols in the above figure can be described by the following formulas:

$$s_t = \tanh(Ux_{t-1} + Ws_{t-1}) \qquad (2-1)$$

$$o_t = \text{softmax}(Vs_t) \qquad (2-2)$$

$$E_t = -\hat{o}_t \log o_t \qquad (2-3)$$

$$E = \sum_t^T E_t \qquad (2-4)$$

As we can see from the above figure and formulas, there are hidden states labelled as $s_t$ (t=1, 2, ⋯t-1, t, t+1 ⋯) in the unfolded RNN cell. The hidden state is the special unit that is used by the RNN. It is a nonlinear transformation unit as mentioned previously and can be obtained from formula (2-1). When t=0, $s_{t-1}$ will be all 0s. Through this operation, hidden states extract all previous input data features and map them into a higher dimension; therefore the RNN can record the data features extracted from previous input data [10]. This is how the memory feature of an RNN arise.

The other important feature of the RNN is processing of sequence data. The $x_t$ (t=1, 2, ⋯t-1, t, t+1 ⋯) in figure 2-3 represents the input vector arranged in time sequence. Each input vector combined with the previous hidden state generates the input for the next hidden state unit. This is the reason for the name "recurrent neural network". Therefore the RNN is suitable to process natural language and other time sequence prediction events [11]. In these events, the input data can be received as sequence data, e.g. word by word in natural language processing, or frame by frame in automatic spoken language recognition.

The cross entropy is often used to calculate the loss between the predicted value and the target value [12]. In supervised learning, the loss function is used to measure the difference between prediction values and actual value. As the formula (2-3) shows, the cross entropy $E_t$ reflects the loss in step t, and the formula (2-4) is the accumulation of loss in the whole sequence T [13]. This feature will make the back propagation more complicated compared to a conventional neural network, because every time step has to be considered when calculating the gradient descent. Therefore the Back Propagation Through Time (BPTT) [14] algorithm is used to upgrade the weights of the RNN.

Compared to a conventional neural network, an RNN has shared weights for the same hidden layer. In the unfolded RNN cell, each step shares the same weights (W, V, U); the only difference is that they have different inputs. The shared weights feature reduces the number of weights needed, thus giving a saving when training a neural network model.

For solving the gradient vanishing and explosion problem in a vanilla RNN [15], researchers initiated the Long Short Term Memory (LSTM) network [16], which is also a kind of RNN. But there are some differences in the basic unit. The basic unit of the LSTM is shown in Fig.4. Compared to the "vanilla" neural network, we can see that there are more cells in the recurrent modules. These cells serve different functions. Basically, there are three kinds of gates in the LSTM cell. They are the input gate, the forget gate, and the output gate.

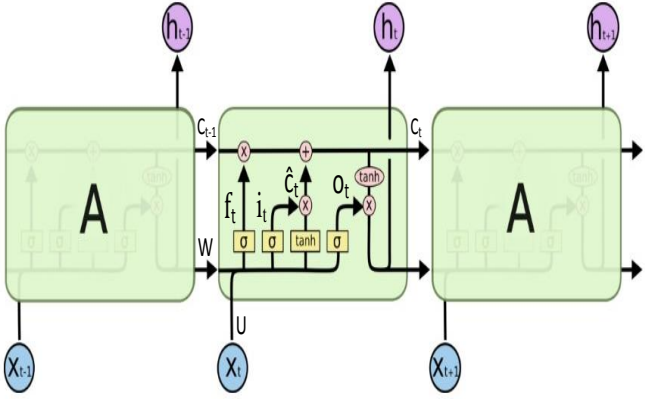

Fig.4. LSTM basic unit modified from [22]

As their names indicate, the input gate decides which kinds of information goes through into the next cell, the forget gate controls the information which should be forgotten, and the output gate finally outputs the previous information to the next cell. The formulas for these symbols are listed as follows:

$$f_t = \sigma(W_f h_{t-1}, + U_f x_t) + b_f \qquad (2-5)$$

$$i_t = \sigma(W_i h_{t-1}, + U_i x_t) + b_i \qquad (2-6)$$

$$\widetilde{C}_t = \tanh(W_g h_{t-1}, + U_g x_t) + b_g \qquad (2-7)$$

$$C_t = f_t * C_{t-1} + i_t * \widetilde{C}_t \qquad (2-8)$$

$$o_t = \sigma(W_o h_{t-1}, + U_o x_t) + b_o \qquad (2-9)$$

$$h_t = o_t * \tanh(C_t) \qquad (2-10)$$

The reason that LSTM can avoid gradient vanishing is the way it calculates the inner hidden states. In a conventional RNN, the inner hidden states can be represented as in formula 2-1. According to the chain rule, the gradient of the hidden state in the back propagation sequence can be calculated by the product of the gradient in every time step [17]. In this case, the product of the gradient will might eventually come to zero due to containing many gradients whose values are far less than 1.

However, things are different in an LSTM. As we can see from the above formulas, the main difference of the LSTM from the general RNN is in the way of updating the hidden state $C_t$. In the data flow path from the previous hidden state $C_{t-1}$ to the next hidden state $C_t$, there is a multiply unit and an addition unit while there is only a multiply unit in the conventional RNN cell. The addition unit makes the back propagation gradient larger than 0 at all times.

### B. Language Model

Language Model (LM) which is involved in machine translation, voice search, speech tagging, and speech recognition, is a fundamental topic in the Natural Language Processing (NLP) field. The role of the LM is to predict the probability of a sentence. LM can be represented as the probability of a word sequence [18]. Its mathematical expression is listed as follows.

$$P(w_1, \ldots, w_T) = \prod_{i=1}^{T} P(w_i | w_1, \ldots, w_{i-1}) \qquad (3-1)$$

$P(w_1, \ldots, w_T)$ represents the probability of the word sequence $(w_1, \ldots, w_T)$. $w_i$ is the i[th] word in the sequence. This formula means that the probability of a sentence is equal to the multiplication of the conditional probability of each word in the sequence, and the conditional probability of each word is determined by all its previous words. This probability model is called an n-gram probability model (the probability of $w_i$ is determined by its n-1 previous words). Specifically, the more previous words that are involved, the more accurate results we could expect. However, it needs a large memory to store these probability weights. For example, if the vocabulary size is 10000, the parameters' volume will be $10000^n$. Meanwhile, we need more training date if we use a larger n. Practically, n is set to 2 or 3 to give a balance between model performance and parameter volume.

Neural Network Language Model (NNLM) [19] was first initiated by Bengio et al at 2003. The NNLM uses a neural network to calculate the probability of sentences, pointing the way to natural language processing. Its system architecture is shown as follows. As can be seen from the diagram, there are three parts: feature vectors input, nonlinear transform, and softmax output, which is the basic structure of the neural network.

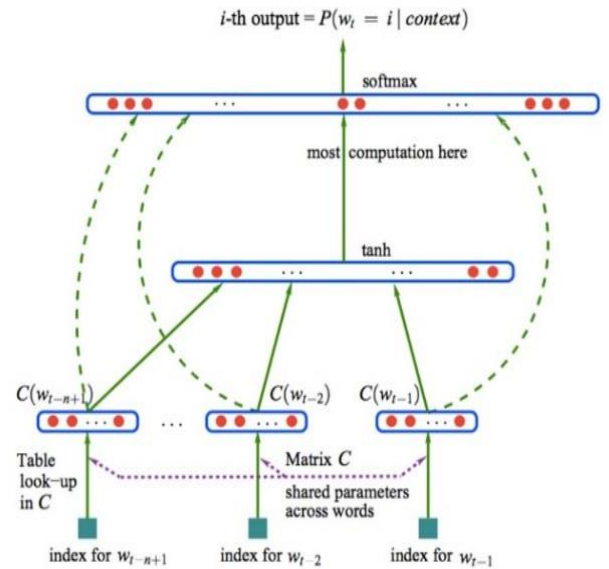

Fig.5. NNLM system diagram modified from [19]

### IV. IMPLEMENTATION

#### A. Software libraries and versions

On the PC side, the main software packages used include Vivado Design Suite 16.1, Anaconda, Python, Theano, and NLTK. We used the Vivado tool to design the Overlay and RTL code running on the PYNQ. We use Conda 4.3.22 on an X86 CPU platform as the software management environment because anaconda software provides an exc ellent scheme to manage software libraries. The Python version is 3.4.5, which is the same version with that in PYNQ board. The Theano version is 0.8.2, which is used to construct mathematic operations. The NLTK version is 3.0.5, which is used to process natural language and build the interface between the corpus dataset and the neural network in Python programs.

On the PYNQ side, the image version is pynq_z1_image_2016_09_14, which includes Python and other software libraries. The original Python version in the system was 2.7. We might need to upgrade it to version 3.4.5. The Jupyter notebook version is 4.0.4.

*B. PYNQ board configuration*

In this section, we will briefly introduce how to boot the PYNQ board and install dependent software packages through the Internet. The pynq_z1_image_2016_09_14 image file includes only Python2.7, Jupyter notebook, and a few other software packages. We needed to install or upgrade the necessary packages by ourselves.

We can download the image file from http://www.pynq.io, and then decompress it into a micro sd card larger than 8G. Finally, insert the micro sd card into sd card slot in the PYNQ board, connect power to the board, and turn on the power switch.

In practice, the best way to access to internet for PYNQ is to use the wireless usb wifi dongle, which is a tiny wifi adaptor with an usb interface. In order to use the wifi connector, we need to download the usb wifi driver file from the PYNQ github: https://github.com/Xilinx/PYNQ, the usb wifi driver file name is Usb_Wifi.py. Copy the driver file into \\pynq\xilinx\pynq\drivers on PYNQ board. The base.bit file, which contains PS and PL design, needed to be downloaded through Overlay interface, and then combined with the driver, the usb wifi connector will work.

*C. DRNN Language Model Design and Deployment*

In this section, we will concretely describe the DRNN Language Model design, training, and deployment methods and procedures. Because the training of the neural network is a time and computing resource consuming work, we use a CPU to train the network, presenting and deploying the model on the PYNQ board.

The DRNN LM program includes 3 principal parts: preprocessing the dataset, building the DRNN model, and training the model. The preprocessing dataset module generates training data from a corpus, converting the words into vectors which can be calculated by the computer. The DRNN model module defines the DRNN topology and algorithms. The training module combines the above modules together to perform the training process and generates prediction results. The DRNN LM program flow chart is listed in Fig.6.

Before going in to concrete module design, we will first present the DRNN LM system architecture. As mentioned at section 3.1, an RNN is suitable for processing of sequence data. We illustrate the data processing in Fig.7.

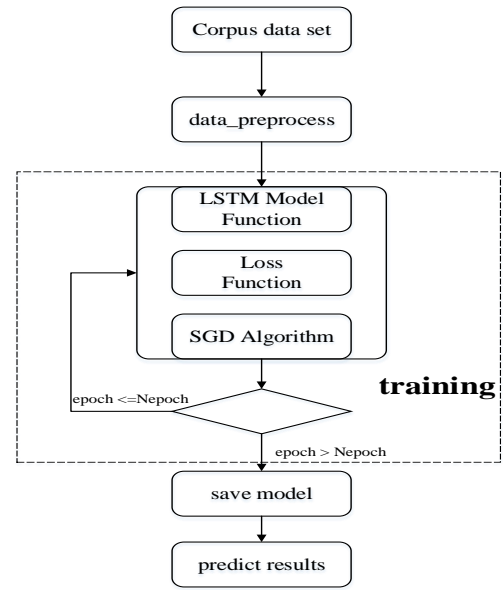

Fig.6. DRNN LM Program Flow Chart

The DRNN LM receives a word sequence input and generates a prediction sequence with the same length. There may be a gap between the prediction sequence and the label sequence as the red marked words in Fig.7, which will result in an increase in the value of the loss function. Then the DRNN weights will be adjusted according to the value of the loss. That is the meaning of the training of the neural network, reducing the loss values and making the prediction close to the label. In the training process, we don't need any pre-generated label files. We just use the dataset in text form and generates the label sentences by the data pre-processing module.

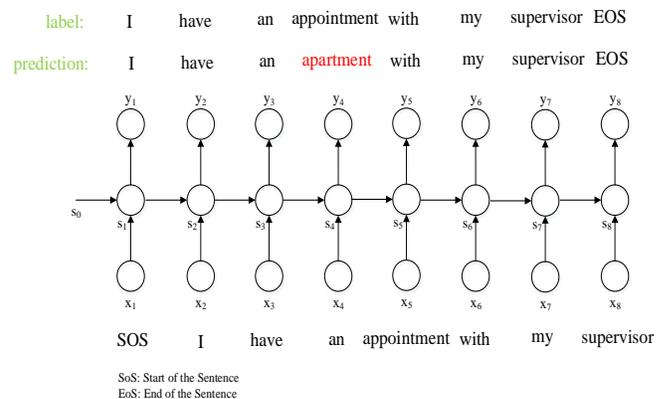

Fig.7. RNN LM system diagram

The first step is to translate the words in human natural language into vectors which can be calculated by the computer. Neural network algorithms use arithmetic operations such as multiplication or addition. So to be suitable for neural network processing, the computer or processor has to translate the symbols into binary or other digital numbers. In this stage the NLTK [20] software package is used to process the natural language dataset, converting the sentences into separated words. After the separated words have been obtained, they must be converted to digital numbers. The Python language has a list variable that can store a word and its corresponding digital number. The words in the dataset are sorted according to frequency of occurrence. Finally, based on the vocabulary dataset, we can get a vocabulary list where each word has a unique one-hot vector coding.

The next step is to build a LSTM Neural Network. Basically, there are two stages in designing a neural network [23]: forward process and backward process. In the forward process, we need to define the neural network topology and neural cells. We use 3 LSTM hidden layers neural network to perform our task. The topology of 3 LSTM hidden layers neural network is shown as follows.

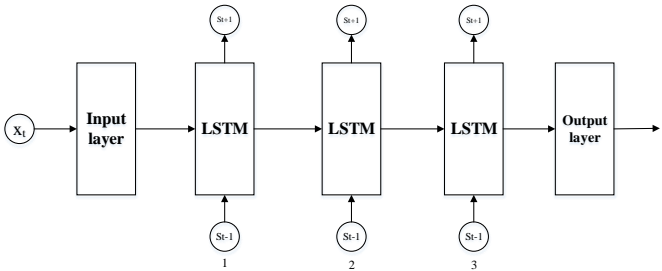

Fig.8. 3 LSTM hidden layers neural network topology

The forward process will transform the data from input to output according to the formulas (2-5)—(2-10). The following expressions in the below figure show the relation between the Python code and the corresponding formulas.

```
f_t1 = T.nnet.hard_sigmoid(U[1].dot(x_e) + W[1].dot(s_t1_prev) + b[1])     →   f_t = σ(W_f · [h_{t-1}, x_t]) + b_f)
i_t1 = T.nnet.hard_sigmoid(U[0].dot(x_e) + W[0].dot(s_t1_prev) + b[0])     →   i_t = σ(W_i · [h_{t-1}, x_t]) + b_i)
ĉ_t1 = T.tanh(U[3].dot(x_e) + W[3].dot(s_t1_prev) + b[3])                  →   C̃_t = tanh(W_C · [h_{t-1}, x_t] + b_C)
c_t1 = f_t1 * c_t1_prev + ĉ_t1 * i_t1                                       →   C_t = f_t * C_{t-1} + i_t * C̃_t
o_t1 = T.nnet.hard_sigmoid(U[2].dot(x_e) + W[2].dot(s_t1_prev) + b[2])     →   o_t = σ(W_o · [h_{t-1}, x_t]) + b_o)
h_t1 = T.tanh(c_t1) * o_t1                                                  →   h_t = o_t * tanh(C_t)
```

Fig.9. LSTM forward process

Because the hard_sigmoid function is faster than a standard sigmoid function, we use hard_sigmoid function to perform the sigmoid operation. The difference of hard_sigmoid and sigmoid function can be seen from the following figure.

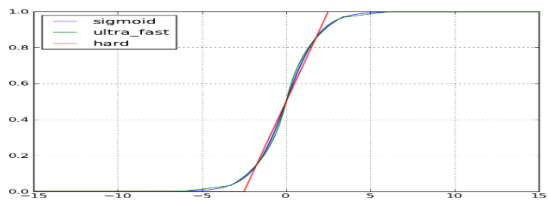

Fig.10. the difference of hard_sigmoid and sigmoid function

The final stage of the forward propagation is the output layer. We use the softmax function in the output layer, which is a kind of normalization function that can convert a vector into a probability distribution form, giving every element a probability. This method is widely used in classification tasks in neural networks. In Theano, we can call softmax function directly. The definition of softmax is described as (5-1) shown and expression (5-2) shows the Theano implementation of the softmax function with Python.

$$\sigma(z)_j = \frac{e^{z_j}}{\sum_{k=1}^{K} e^{z_k}} \qquad j \ in \ 1,\dots,K \qquad (4-1)$$

$$T.nnet.softmax(V.dot(s)) \qquad (4-2)$$

Now we need a loss function to calculate the gap between the prediction and the actual values, which will be used to get the gradient of every weight in the back propagation. Because the cross entropy loss function is faster in training than the mean square error function, we use a cross entropy loss function whose definition is shown as (5-3), and expression (5-4) shows the Theano implementation of cross entropy with Python.

$$H(p.q) = -\sum_{i=1}^{n} p(x) \log(q(x)) \qquad (4-3)$$

$$T.nnet.categorical\_crossentropy(o, y) \qquad (4-4)$$

The backward process of the h neural network is to calculate the gradients of the neural network, using the Back Propagation Through Time (BPTT) algorithm [24]. In practice, Theano supports auto-differentiation of formulas. We can get the gradients through the grad function in Theano.

The neural network needed to be trained for hundreds of epochs to make the prediction values close to the actual values. One epoch includes one forward propagation and one backward propagation. We use the Stochastic Gradient Descent (SGD) algorithm to update the weights of the neural network through calculating the partial derivatives of the loss function with respect to weights and biases.

In terms of evaluating the language model, we need to check the loss values and the perplexity of the language model. The two indexes are simple and fast to calculate and can make the performance of the model readily comprehensible [25]. We have introduced the cross-entropy loss function above. In this section, we will mainly introduce perplexity. Perplexity is a kind of intrinsic evaluation method for a language model [26]. It can be obtained through taking the reciprocal of the probability of a test sentence and then computing the Nth root [27]. If taking the logarithm, in practice, the perplexity can be expressed as cross-entropy to the power of e. More specifically, the perplexity means in a specific context, the number of words that the language model might choose as the next word. Therefore, the smaller perplexity, the higher probability of a sentence, and the better model. In Python, it can be calculated as the following expression:

$$np.\exp(np.mean(total_{loss}/iterations)) \qquad (4-5)$$

In Theano, the trained neural network model is saved in a *.npz file, which saves several arrays into a single file in uncompressed format. The final size of the model is around 17MB, which can be easily deployed into the PYNQ board. When we get the model of the LSTM LM neural network, we can use it to predict results on the PYNQ. The model file needed to be copied to the PYNQ board file system, and loaded with the following instructions in Jupyter notebook. Then we can use the model on PYNQ board.

```
# Load parameters of pre-trained model
model = load_model_parameters_theano('./data/lstmrnn.npz')

Building model model from ./data/pretrained.npz with hidden_dim=128 word_dim=8000
```

Fig.11. model load instructions in Jupyter notebook

### D. DRNN Accelerator Design

An overlay is built in programmable logic on the ZYNQ system [28]. It is a kind of hardware library which has a Python interface, making it convenient for the software programmer to use overlays as Python packages. Therefore, the overlay includes a PL part and a Python interface part. We can design the PL part and generate the corresponding tcl file with the Vivado Design Suite tool. The tcl file for the PL design can be automatically identified by the PYNQ system and used as an interface connecting to Python methods [28]. In order to use the overlay, we need to copy the bit and tcl files, which are generated by Vivado Design Suite with the command: $write\_bd\_tcl$, to the Python file system.

Vivado Design Suite 16.1 is used to create a new overlay through building a new block design and adding the new IP modules. The PYNQ project in GitHub [29] has provided a base design which includes most of the peripheral interface overlay. Therefore, we don't need to build everything from scratch. We will build the accelerator overlay based on this project. In order to reuse the base PYNQ project, we create a Vivado project in .\vivado\base, and import the tcl file and bitstream file. The Tcl Console windows is in the bottom of the Vivado Design Suite tool, and the command is "source …/***.tcl".

After importing the base PYNQ block design, we start to add the accelerator overlay into it. The accelerator overlay needs an AXI Stream interface to communicate with the PS, because we transfer data as batches, and the AXI Stream performs data batch transferring very well. In Vivado Design Suite, an IP with AXI Stream interface can be created through going to Tools/Create and Package IP option. Finally, we generate an AXI Stream interface IP, add it into the Vivado IP repository automatically, and connect it to the PS part in the block design. The block diagram of the accelerator overlay data path in block design is shown in Fig.12.

Because AXI Stream cannot directly map its data into memory address on the PS side, the AXI DMA modules are used to map the data from FIFO to the memory address, and vice versa. The acceleration procedure is: the accelerator IP receives a batch of data from the PS, finishes the acceleration operations, and sends the results to the PS.

The DRNN accelerator system is implemented with Verilog Hardware Language on the PL side of the ZYNQ SOC. In order to improve the parallel computation level, the accelerator core has 5 parallel Processing Elements (PE). The accelerator system architecture is illustrated as follows fig.13.

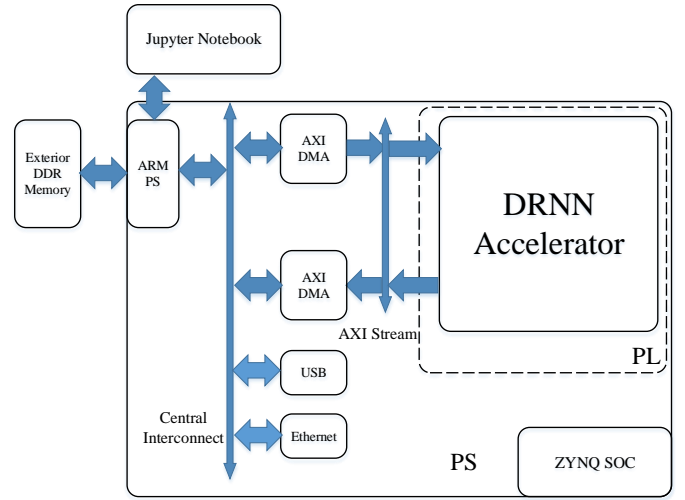

Fig.12. accelerator overlay data path diagram

The accelerator performs matrix multiplication operations which are the most frequent operation in forward propagation of the neural network. We design a multiplication and addition accelerator which accelerates $W_f h_{t-1} + U_f x_t$ operation in a language model, where $W_f \in R^{50 \times 50}$, $h_{t-1} \in R^{50 \times 1}$, $U_f \in R^{4000 \times 50}$, $x_t \in R^{4000 \times 1}$ (Vocabulary size = 4000, hidden layer cell = 50). There are five Process Elements (PE) implemented in one accelerator core unit. Each PE receives 50 data in one batch and performs 50*10 times multiplication operations and 50*10 times addition operations.

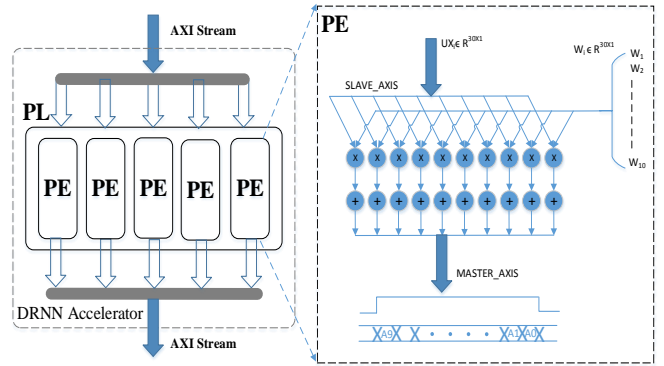

Fig.13. multiplication and addition accelerator basic unit and hardware implement

The whole accelerator will perform 50*10*5 times multiplication operations and 50*10*5 times addition operations in one batch. The latency of the accelerator is around 50*5ns. That means the number of operations that the accelerator executes per second is around 20GOPS (measured in Operations per Second). Because all hidden layer hold the same weights, the accelerator overlay can be reused by all of them.

## V. EXPERIMENT

We use Jupyter notebook and Vivado Design Suite to build a joint simulation system. Using overlay, a Python program can control the internal DMA IP address to launch data transmission, and external data sends/receives to/from the accelerator through Python API and AXI interface. The joint simulation system diagram is shown as follow.

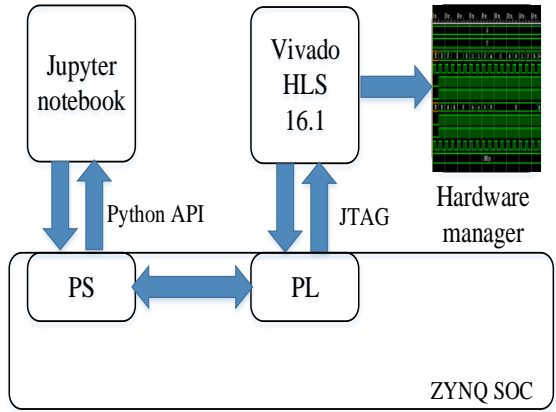

Fig.14. joint simulation system framework

We send consecutive numbers from 1 to 50 into the hardware accelerator, and get the results back to the program.

The hardware calculation results by hardware are:

*sum_hardware=[1275,2550,3825,5100,6375,7650,8925,10200,11475,12750, 14025,15300,16575,17850,19125,20400,21675,22950,24225, 25500,26775,28050,29325,30600,31875,33150,34425,35700, 36975,38250,39525,40800,42075,43350,44625,45900,47175, 48450,49725,51000,52275,53550,54825,56100,57375,58650, 59925,61200,62475,63750]*

Fig.15. hardware calculation results

In order to verify the results of hardware implemention, we calculate the multiplication and addition by using a computer first. The results by computer are as follows.

```
Output:

for j in range(1,51):
    sum =0
    for i in range(1,51):
        sum = i*j + sum

sum =[
1275,2550,3825,5100,6375,7650,8925,10200,11475,12750,14025,15300,16575,17850,
19125,20400,21675,22950,24225,25500,26775,28050,29325,30600,31875,33150,34425
35700,36975,38250,39525,40800,42075,43350,44625,45900,47175,48450,49725,51000
52275,53550,54825,56100,57375,58650,59925,61200,62475,63750]
```

Fig.16. the multiplication and addition by using computer

We use matplot library to display the above accumulation results by hardware and software. It can be seen from the Fig.21, the results by hardware match those by software very closely.

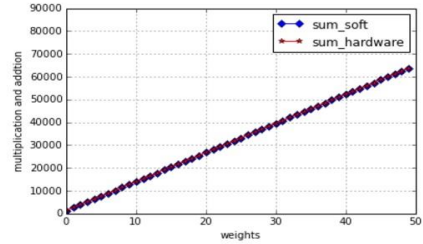

Fig.17. the accumulation results by hardware and software

## VI. RESULTS

### A. Cross-Entropy Loss Function Results

We obtained the loss values of 245 epochs in the training processing and plot them in Jupyter notebook with the matplotlib package. As can be seen from the figure below, the loss values are gradually reducing with the increase of training epochs, which means the training algorithms works well and the prediction of the model will be close to the label when the training process is finished.

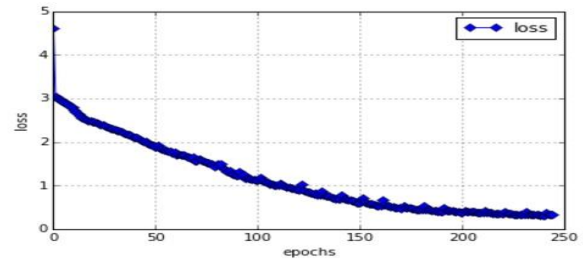

Fig.18. the loss values of 245 epochs

### B. Language Model Perplexity

We obtain the perplexity in every 100 training steps. The result is shown in Fig. 23. The perplexity decreases from around 3500 to about 100 after 1000 training steps, which means the language model is reducing the scope of vocabulary pool when choosing the next words to predict. That means the .prediction speed and accuracy rate of the language model is increasing with the progress of training.

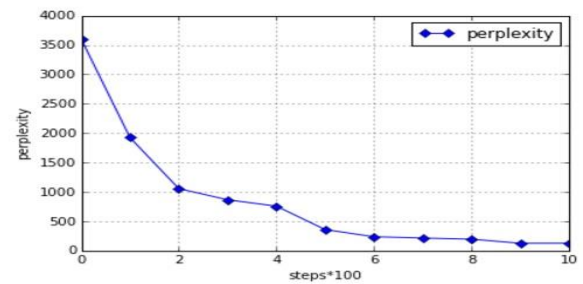

Fig.19. perplexity of language model

## C. DRNN Accelerator Performance

The FPGA resource utilization is listed as follow. We use speed optimization option when generating the multiplier IP core, therefore each multiplier IP core consumes 4 DSPs. We have 50 multipliers in our accelerator module. BRAM resource is used to restore the weights. LUT resource is consumed to register and routing. The overall resource utilization is very well.

Table 1: FPGA resource utilization

| Resource | Utilization | Available | Utilization |
|---|---|---|---|
| LUT | 39851 | 53200 | 74.91% |
| LUTRAM | 3245 | 17400 | 18.65% |
| FF | 54488 | 106400 | 51.21% |
| BRAM | 75.50 | 140 | 53.93% |
| DSP | 206 | 220 | 93.64% |

In order to evaluate the accelerator performance, we compare our implementation to previous related works. The comparison dimensions includes FPGA type, running frequency, precision, DSP utilization, neural network type, and acceleration performance. The work in Ref. [30] implements a whole LSTM cell on FPGA and the acceleration performance is 0.2837GOPS. Ref. [31] implements the NNLM on a larger FPGA which has 2800 DSP resources overall. Compared to the two related works, our implementation has 70.5X and 2.75X speedup respectively.

Table 2: Accelerator performance comparison to related work

| | Ref.[30] | Ref.[31] | Our implement |
|---|---|---|---|
| FPGA type | Zynq 7020 | Virtex7-485t | Zynq 7020 |
| Frequency/MHz | 142 | 150 | 200 |
| Precision | Fixed point(Q8.8) | Float-32 | Fixed point |
| DSP utilization | 50 | 1176 | 206 |
| Neural network | RNN-LSTM | RNN-LSTM | RNN-LSTM |
| Performance | 0.2837GOPS(1X) | 7.26GFLOPS(25.6X) | 20 GOPS(70.5X) |

## VII. CONCLUSION AND FUTURE WORK

The paper implements a DRNN language model with Python and deploys the trained model on PYNQ through Jupyter notebook, and designs an DRNN hardware accelerator using an overlay. We use a high level programming language—Python - to design an FPGA SOC system, which greatly accelerates the development process and extends the range of FPGA application. We build a hardware accelerator with AXI Stream interface, which can interact with a software program through overlay form. The acceleration performance has a great improvement compared to previous related works.

More importantly, we showed that a software and hardware joint design and simulation process can be useful in the neural network field. The model trained through CPU or GPU can be deployed into the FPGA SOC system to explore the application on a mobile device. Meanwhile, the programmable logic on the FPGA can be used as a hardware accelerator to improve the performance of the model. The project is uploaded into https://github.com/hillhao/PYNQ-project .

There are several points that can be optimized or studied further. The word-to-index part of the language model can be optimized to low density vector, which can greatly reduce the magnitude of dimensions of word vectors. The training methods, which can shorten the training length, accelerate convergence, and avoid over-fit, need to be studied further. In the hardware side, implementing all the neural network parts on the FPGA SOC platform would be beneficial because of its low power consumption and powerful parallel processing features.